\documentclass[lettersize,journal]{IEEEtran}
\usepackage{amsmath,amsfonts}
\usepackage{algorithmic}
\usepackage{algorithm}
\usepackage{array}
\usepackage[caption=false,font=normalsize,labelfont=sf,textfont=sf]{subfig}
\usepackage{textcomp}
\usepackage{stfloats}
\usepackage{url}
\usepackage{verbatim}
\usepackage{graphicx}
\usepackage{cite}

\usepackage{booktabs}
\usepackage{multirow}

\usepackage{tikz}
\usetikzlibrary{arrows.meta,positioning}

\usepackage{orcidlink}

\usepackage{authblk}

\begin{document}

\title{TacPrint: A Wearable Fingertip Tactile Sensor \\ for Human-to-Robot Contact Reproduction}

\author[1]{Yongxi Liu}
\author[1]{Chaofan Zhang}
\author[1]{Xingyu Zhang}
\author[1]{Xiangyin Bao}
\author[1]{Boyue Zhang}
\author[1,2,3,*]{Shaowei Cui}
\author[1]{Shuo Wang}

\affil[1]{Institute of Automation, Chinese Academy of Sciences}
\affil[2]{Imprintx Robotics}
\affil[3]{Beijing Academy of Artificial Intelligence}




\maketitle

\begingroup
\renewcommand{\thefootnote}{*}
\footnotetext{Corresponding author:  \texttt{shaowei.cui@ia.ac.cn}}
\endgroup

\begin{abstract}
Human-centric data collection is emerging as a significant paradigm for robot skill acquisition, but seamlessly integrating low-cost, scalable tactile sensing systems that capture fine-grained fingertip interactions without compromising natural operation remains a key challenge. 
This reduces the reliability of human-to-robot transfer in contact-rich tasks.
In this work, we present TacPrint, a wearable fingertip tactile sensor, where protrusions on the inner surface of the silicone skin are aligned one-to-one with 24 capacitive taxels to enable localized capacitive responses.
A real-to-sim-to-real pipeline estimates a $35\times26$ contact-depth map from 24-channel capacitive signals. Against simulation-generated labels, the model achieved a contact-region RMSE of $0.223\pm0.161$~mm, a weighted-centroid error of $1.213\pm2.379$ pixels, and an IoU of $0.829\pm0.169$. With measured capacitive inputs, the network-predicted depth evaluated at the guide-calibrated contact center showed a mean absolute error of $0.085\pm0.057$~mm across all 40 controlled trials, while the mean contact-position error was $0.250\pm0.208$~mm across the 37 trials whose reference contact regions were not truncated by the sensing boundary.
In human-to-robot replay, tactile-guided compensation increased grasping and wiping success rates from 0\% to 91.67\% and 90\%, respectively. In closed-loop grasping, dense-depth feedback achieved success rates of 87.5\% over all tested positions and 85\% under edge-contact conditions, compared with 67.5\% and 45\% for raw-taxel feedback.

\end{abstract}


\section{Introduction}

\IEEEPARstart{H}igh-quality data is central to intelligent robotic manipulation, as the performance of skill learning and transfer is fundamentally bounded by the fidelity of the demonstrations collected. To obtain such data, various collection paradigms have been developed, ranging from teleoperation and kinesthetic teaching~\cite{aloha1,aloha2,gello,HowToTrainYourRobots,LearntoGraspObjwithDexRobotManipulator} to motion-capture-based and other demonstration-driven approaches~\cite{umi,FreeTacMan,dexcap,arcap,dart,DexUMI,DexViTac}. Among them, human-centric data collection has emerged as a promising direction because it enables natural and efficient acquisition of manipulation behaviors, especially for dexterous, contact-rich tasks. However, current human-centric pipelines still predominantly focus on vision and kinematics and therefore lack precise tactile acquisition of the contact states critical to manipulation.

To bridge this gap, a range of tactile sensors has been developed to recover contact geometry and local interaction patterns, including several vision-based or capacitive high-resolution tactile sensing systems~\cite{DIGIT, Insight, Dtact, DotTip, GelStereoPalm2.0, pptac}. These devices demonstrate the ability to capture fine-grained contact information, including shape, deformation, and depth. For human-centric data collection, however, tactile sensing should ideally be embedded in a form factor that can be worn directly on the fingers during natural demonstrations. Motivated by this need, prior studies have explored wearable tactile fingertips~\cite{ThimbleSense, Skin-like, FingerTac, TacCap}, which show that rich contact information can, in principle, be acquired in a wearable manner. However, existing solutions still struggle to simultaneously provide compact wearability, an informative local contact representation, adaptable mounting for human fingertips and selected robotic hands, and low-cost scalable deployment. Achieving a balance between sensing richness and practical usability therefore remains challenging.

\begin{figure}[!t]
\centering
\includegraphics{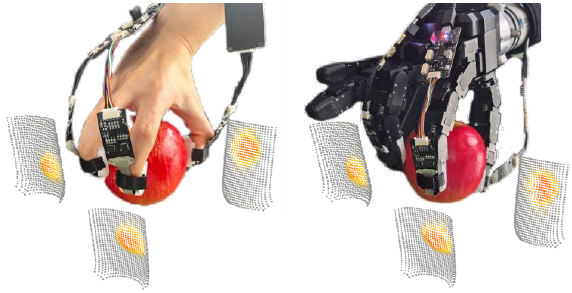}
\caption{TacPrint is a compact, low-cost wearable fingertip sensor that estimates a $35\times26$ contact-depth map from 24-channel capacitive signals.}
\label{S1_intro.pdf}
\end{figure}

We present TacPrint, a compact wearable fingertip sensor that adds 7.8~mm of fingertip thickness and costs approximately \$50. An adjustable Velcro strap accommodates the tested range of human fingertip dimensions, while customizable connectors enable installation on selected robotic hands. Its taxel-aligned silicone protrusions provide localized and structured contact transduction.

To recover dense contact-depth information from sparse measurements, we map short temporal sequences of 24-channel capacitive measurements to $35\times26$ depth maps. We conducted three complementary evaluations: comparing configuration-referenced depth and contact location from simulation-generated labels with controlled physical contacts to examine the label-generation procedure; comparing network predictions from measured capacitive inputs with the corresponding physical contact configurations to assess reconstruction accuracy; and applying the reconstructed contact information obtained during wearable use to robot replay and closed-loop grasp adjustment to evaluate its task-level utility.

The contributions of this study are summarized as follows:
\begin{enumerate}
\item We present TacPrint, a compact and low-cost wearable fingertip sensor with adjustable human mounting, customizable robotic connectors, and taxel-aligned silicone protrusions for localized contact transduction.
\item We develop a tactile-to-depth learning pipeline that estimates a $35\times26$ contact-depth map from 24-channel capacitive signals, and evaluate its depth and contact-location consistency using simulation-generated labels and controlled physical configurations.
\item We demonstrate tactile-guided human-to-robot replay and closed-loop contact-position-aware grasping. Tactile compensation increased grasping and wiping success rates to 91.67\% and 90\%, while dense-depth feedback achieved 87.5\% overall and 85\% edge-contact grasping success, compared with 67.5\% and 45\% for raw-taxel feedback.
\end{enumerate}

\section{Related Works}
\subsection{Data Collection}
Data collection for robotic manipulation has increasingly shifted toward human-centric paradigms that better preserve natural hand-object interaction during demonstrations. Systems such as UMI~\cite{umi} and FreeTacMan~\cite{FreeTacMan} support intuitive data acquisition, but still require the operator to hold handheld grippers. Solutions such as DexCap~\cite{dexcap}, ARCap~\cite{arcap}, and DART~\cite{dart} eliminate the need for external devices and directly capture human fingertip motion, making the collection process more natural. DexUMI~\cite{DexUMI} and DexViTac~\cite{DexViTac} further enrich this setting by acquiring tactile information together with hand motion, improving demonstration fidelity while maintaining the advantages of human-centric collection for skill transfer. Nevertheless, precise acquisition of local fingertip contact states remains insufficient, which motivates the need for wearable tactile sensing in contact-rich manipulation.

\subsection{Wearable Tactile Fingertips}

Wearable tactile sensing at the fingertip has emerged as a promising direction to enrich human-centric manipulation data with local contact information. ThimbleSense~\cite{ThimbleSense} introduced a fingertip-wearable thimble equipped with a force/torque sensor, enabling the measurement of contact forces and estimation of contact locations for grasp analysis. FingerTac~\cite{FingerTac} further explored a more anthropomorphic and interchangeable design by distributing Hall-effect sensing elements over the palmar and lateral fingertip surfaces, allowing both human and robotic hands to capture distributed three-axis tactile signals. TacCap~\cite{TacCap} more recently proposed a lightweight FBG-based tactile thimble for seamless human-to-robot skill transfer, emphasizing portability, durability, and cross-platform deployment on both human and robotic fingertips. 
Together, these works motivate a wearable tactile interface that combines compact structure, informative local sensing, adjustable human mounting, connector-based robotic installation, and low-cost deployment.

\section{Methods}
\begin{figure}[!t]
\centering
\includegraphics[width=3.5in]{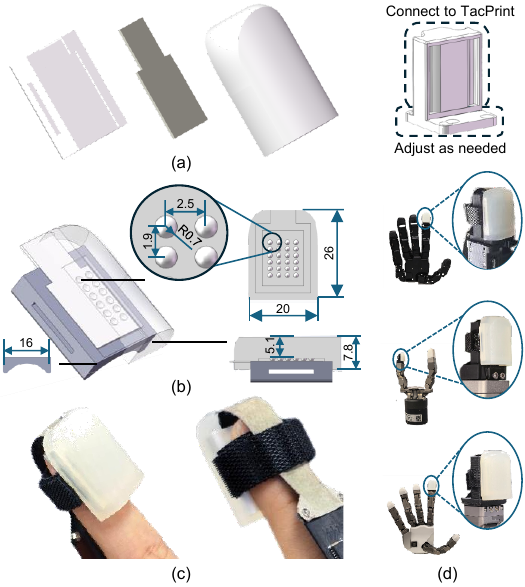}
\caption{Structure design of TacPrint. (a) TacPrint components: a silicone skin, a 24-channel capacitive sensing unit, and a 3D-printed support. (b) Size parameters of the silicone skin. (c) Two views of TacPrint worn on a human fingertip. (d) Connector and representative TacPrint installations on three robotic hands. The connector can be customized for the target mounting interface.}
\label{S3_structure}
\end{figure}

\subsection{Structure Design} 

TacPrint comprises a silicone skin, a 24-channel capacitive sensing unit, and a 3D-printed support, as shown in Fig.~\ref{S3_structure}(a). The silicone skin, made of Mold Star 20T with a Shore hardness of 20A, contains a $6\times4$ array of 24 hemispherical protrusions aligned one-to-one with the capacitive taxels~\cite{DotTip}. Each protrusion has a radius of 0.7~mm, with row and column spacings of 1.9 and 2.5~mm, respectively, as shown in Fig.~\ref{S3_structure}(b). The silicone was fabricated by mixing Parts A and B at a 1:1 ratio by weight, vacuum-degassing the mixture, and curing it at room temperature for 40~min. The mold and support had a nominal printing accuracy of 0.1~mm; the corresponding mold--support dimensional mismatch is conservatively bounded at 0.2~mm, while bonding-induced variation was not separately quantified. The capacitive sensing unit is the 24-channel standard edition of Model~2 from SNIOW Technology.

An adjustable Velcro strap, shown in Fig.~\ref{S3_structure}(c), enables rapid attachment without imposing a predefined cavity size. Across 12 participants with fingertip widths of 14--18~mm and circumferences of 45--58~mm, the mean stability and comfort scores were 4.76 and 4.87 out of 5, respectively. Across 57 repeated triangular-indenter trials, centroid-aligned IoU yielded participant-level means of 0.827--0.904 and an overall participant-level mean of $0.861\pm0.022$. No pronounced participant-dependent degradation was observed within the tested range.

For robotic installation, the connector combines a TacPrint-matching upper section with a customizable lower mounting interface, while a rear Velcro strap provides additional retention. Representative installations on the LinkerHand O20, Tesollo DG-2F-M, and Tesollo DG-5F-S are shown in Fig.~\ref{S3_structure}(d); installation requires sufficient fingertip space and a suitable attachment interface.

\subsection{Tactile-to-Depth Learning Pipeline}
\begin{figure}[!t]
\centering
\includegraphics{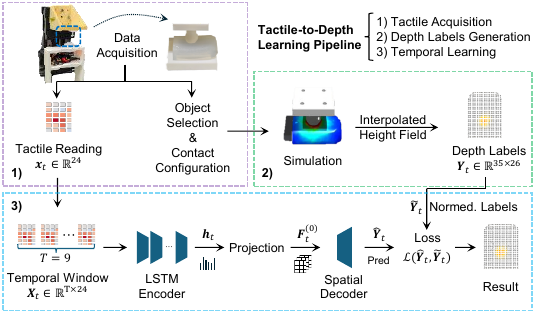}
\caption{Overview of the tactile-to-depth learning pipeline. Controlled physical indentations provide synchronized 24-channel capacitive sequences; the matched nominal contact configurations are reproduced in TacFlex/Isaac Gym/Flex to generate $35\times26$ depth labels; and an LSTM-based temporal encoder and spatial decoder estimate the dense contact-depth map.}
\label{S3_pipeline}
\end{figure}

The proposed pipeline estimates a dense fingertip depth map from TacPrint's low-dimensional tactile response, as shown in Fig.~\ref{S3_pipeline}. At time step $t$, a fixed-length tactile window $\mathbf{X}_t\in\mathbb{R}^{T\times24}$ is mapped to a predicted depth map $\hat{\mathbf{Y}}_t\in\mathbb{R}^{35\times26}$:
\begin{equation}
\hat{\mathbf{Y}}_t=\mathcal{F}(\mathbf{X}_t;\Theta),
\end{equation}
where $\Theta$ denotes the trainable parameters. The pipeline comprises tactile acquisition, simulation-based label generation, and temporal learning.

\subsubsection{Tactile Acquisition}

TacPrint is fixed relative to an XYZ computer numerical control (CNC) linear guide, and the coordinate frame of its active sensing area is registered to the guide coordinate frame before data collection. The CAD geometry and mounting orientation of each rigid indenter define its geometry and orientation relative to the sensor. The unindented state is used as the nominal zero-depth reference, the prescribed guide displacement along the indentation direction defines the nominal indentation depth, and the registered in-plane guide coordinates define the nominal contact location. These quantities jointly specify the nominal contact configuration associated with each measured 24-channel capacitive sequence. During each episode, the guide executes an approach, pressing, holding, and release sequence. The synchronized measurements are subsequently organized into fixed-length inputs for temporal learning.

\subsubsection{Simulation-Generated Depth Labels}

Simulation-generated depth labels are obtained by reproducing each nominal contact configuration, defined by the corresponding indenter geometry, relative pose, contact location, and indentation depth. The pipeline is based on TacFlex~\cite{Tacflex}, with Isaac Gym/Flex as the physics backend. The measured capacitive sequence and simulated label are paired frame by frame according to this prescribed configuration. Here, calibration denotes configuration-level correspondence between the physical and simulated contacts rather than independent full-field deformation calibration.

The TacPrint elastomer is discretized using a tetrahedral volumetric mesh and modeled as a Neo-Hookean material with $E=0.30$~MPa and $\nu=0.47$. The indenter--elastomer friction coefficient is set to 0.20. These parameters are fixed for all simulated contacts and are treated as literature- and benchmark-based effective parameters rather than batch-specific constitutive parameters identified from the fabricated silicone skin. Uncertainty in the material parameters, friction, boundary constraints, fabrication, and bonding may therefore affect the simulated contact boundary and peripheral deformation. Accordingly, the physical consistency of the simulation-generated labels is further examined using controlled contact experiments.

The simulated deformation field is rasterized into a $35\times26$ depth map, providing dense supervision constrained by the measured contact configuration. Configuration-level consistency was evaluated using 78 depth trials and 40 contact-position trials. The mean contact-center depth error was $0.110\pm0.016$~mm. Across all 40 position trials, the mean errors were $0.338\pm0.227$, $0.279\pm0.201$, and $0.265\pm0.186$~mm at segmentation thresholds of 0.3, 0.4, and 0.5~mm, respectively. Although 0.5~mm gave a slightly lower mean error, 0.4~mm yielded lower median and 95th-percentile errors and was therefore used as the representative threshold. At the representative 0.4-mm threshold, the error was additionally evaluated on the 37 trials whose simulated reference contact regions were not truncated by the sensing boundary, yielding $0.247\pm0.159$~mm. These results evaluate the configuration-level consistency of the simulation-generated labels rather than the trained network.

Before training, each depth map is clipped to the valid indentation range and normalized using a fixed global scale:
\begin{equation}
\tilde{\mathbf{Y}}_t=
\frac{\mathrm{clip}(\mathbf{Y}_t,0,D_{\max})}{D_{\max}},
\end{equation}
where $D_{\max}=3$~mm. Unlike sample-wise normalization, this fixed scaling preserves the relative depth magnitudes across different contacts.

\subsubsection{Temporal Learning}

Let $\mathbf{x}_n\in\mathbb{R}^{24}$ denote the capacitive measurement at frame $n$. For an odd window length $T=2K+1$, the input
$\mathbf{X}_t\in\mathbb{R}^{T\times24}$ centered at frame $t$ is
\begin{equation}
\begin{aligned}
\mathbf{X}_t
&=
\left[
\mathbf{x}_{\pi(t-K)},\ldots,
\mathbf{x}_{\pi(t+K)}
\right]^{\mathsf T},\\
\pi(n)
&=
\min\!\left(\max(n,1),N\right).
\end{aligned}
\label{eq:temporal_window}
\end{equation}
where $N$ is the episode length and $\pi(\cdot)$ replicates boundary frames. We use $T=9$ and $K=4$.

The sequence is encoded by a single-layer unidirectional LSTM:
\begin{equation}
\begin{aligned}
(\mathbf{h}_{\tau},\mathbf{c}_{\tau})
&=
\operatorname{LSTMCell}
\big(
\mathbf{X}_t[\tau,:],\mathbf{h}_{\tau-1},
\mathbf{c}_{\tau-1}
\big),\\
\mathbf{z}_t
&=
\mathbf{h}_T.
\end{aligned}
\label{eq:lstm_encoding}
\end{equation}
where the hidden dimension is 256 and $\mathbf{z}_t$ represents the local temporal context of the target frame. The encoded feature is mapped to the depth prediction by
\begin{equation}
\hat{\mathbf{Y}}_t=\mathcal{D}_{\theta}(\mathbf{z}_t),
\qquad
\hat{\mathbf{Y}}_t\in\mathbb{R}^{35\times26},
\label{eq:depth_decoder}
\end{equation}
where $\mathcal{D}_{\theta}$ consists of a linear projection, reshaping to a $32\times9\times7$ feature map, two transposed-convolution layers, and convolutional refinement.

Training uses a foreground-weighted combination of squared and absolute errors:
\begin{equation}
\begin{aligned}
e_{tij}
&=
\hat{Y}_t(i,j)-\tilde{Y}_t(i,j),\\
w_{tij}
&=
1+(\gamma-1)
\mathbf{1}_{\{\tilde{Y}_t(i,j)>\delta\}},\\
\mathcal{L}_t
&=
\frac{1}{HW}
\sum_{i,j}
w_{tij}
\left[
\lambda e_{tij}^{2}
+(1-\lambda)|e_{tij}|
\right].
\end{aligned}
\label{eq:weighted_loss}
\end{equation}
Here, $\mathbf{1}_{\{\cdot\}}$ is the indicator function, and the
summation is taken over all $H\times W$ pixels. We set $\lambda=0.5$, $\delta=0.3$, and $\gamma=2$. The network is trained for 50 epochs using AdamW with a learning rate of $10^{-3}$, weight decay of $10^{-4}$, and batch size of 32.

\section{Experiments}
\subsection{Design and Setup}
The primary goal of our experiments is to evaluate TacPrint at three levels: contact-depth estimation, tactile-informed human-to-robot replay, and the closed-loop use of spatial contact information. Specifically, we aim to answer the following questions:

\textbf{Q1:} {Can TacPrint estimate informative contact-depth representations from sparse capacitive measurements with consistent depth and location under controlled physical indentation?}

\textbf{Q2:} {Can tactile information collected by TacPrint during human demonstrations improve downstream robot replay in contact-rich manipulation tasks?}

\textbf{Q3:} Does the estimated dense-depth representation provide more reliable contact-position feedback for robotic grasp adjustment than contact-only sensing or direct raw-taxel localization?

To answer these questions, experiments are divided into three parts. \textbf{Exp.~1} evaluates the contact characterization capability of TacPrint. \textbf{Exp.~2} evaluates its utility in tactile-informed human-centric data collection and robot replay. \textbf{Exp.~3} evaluates the contribution of the estimated dense-depth representation to closed-loop contact-position adjustment and grasping.

\textbf{Exp.~1: Contact characterization.}

Exp.~1 combines prediction-to-label evaluation, configuration-referenced physical evaluation, and a qualitative continuous-contact demonstration. Predictions on the test set are compared with simulation-generated labels in depth, contact location, and contact-region overlap. In the complementary physical evaluation, the same 40 controlled indentation trials are used to evaluate two quantities. First, the predicted depth at the guide-calibrated contact center is obtained by bilinear interpolation and compared with the nominal indentation depth in all 40 trials. Second, contact position is estimated from the geometric centroid of the predicted region above the 0.4-mm threshold. The primary position statistic includes 37 trials whose reference contact regions remain fully within the sensing area. Three trials are excluded from this statistic because their reference contact regions touch the sensing boundary and are therefore spatially truncated. This exclusion is determined from the reference regions rather than the network predictions.

\begin{figure}[!t]
\centering
\includegraphics{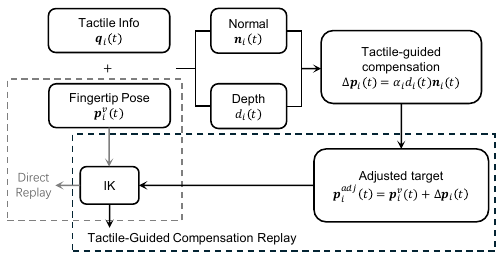}
\caption{Pipeline for tactile-guided compensation. Direct replay uses vision-captured fingertip positions for IK solving, whereas tactile-guided compensation replay uses tactile-adjusted fingertip positions.}
\label{S4_exp2_compensate_pipeline}
\end{figure}

\textbf{Exp.~2: Human-centric data collection and robot replay.}

Exp.~2 includes grasping an apple, mango, and peach, as well as holding an eraser during two back-and-forth whiteboard-wiping passes. TacPrint is worn on the operator's thumb, index finger, and middle finger. A small cube is mounted above each instrumented fingertip, with AprilTags attached to five outward-facing surfaces so that the fingertip pose can be recovered whenever at least one tag is visible. The demonstrations are replayed using a RealMan RM65-B arm and a LinkerHand O20, with TacPrint providing normal-direction tactile compensation. Direct and tactile-compensated replay are compared using task success rate.

\textbf{Exp.~3: Contact-position-aware grasping.}

\begin{figure}[!t]
\centering
\includegraphics[width=\columnwidth]{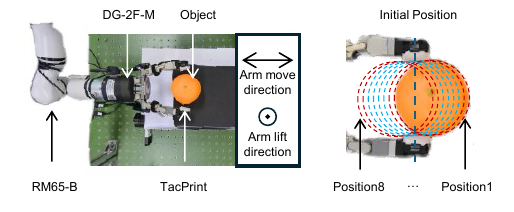}
\caption{Setup of the contact-position-aware grasping experiment. (a) Robotic platform equipped with two TacPrint sensors, with the lateral adjustment and lifting directions indicated by arrows. (b) Eight initial object positions relative to the two TacPrint sensors, ordered from Pos.~1 to Pos.~8 from right to left.}
\label{S4_exp3_setup}
\end{figure}

As shown in Fig.~\ref{S4_exp3_setup}, a RealMan RM65-B arm and a Tesollo DG-2F-M gripper equipped with two TacPrint sensors grasp an orange model placed at eight lateral positions. Each position is tested five times under each feedback strategy, yielding 40 trials per strategy. Pos.~1, Pos.~2, Pos.~7, and Pos.~8 are additionally analyzed as edge-contact conditions, yielding 20 edge-contact trials per strategy.

Three feedback strategies are compared using a quasi-spherical orange model. For this object, the predefined $\pm2$~mm grasping band represents contact near its largest-diameter section, where opposing contacts provide more stable support during lifting. \textit{Contact only} triggers grasping and lifting once valid contact is detected, without lateral adjustment. \textit{Raw-taxel control} estimates the contact center from the normalized $6\times4$ taxel array, whereas \textit{dense-depth control} extracts pixels above 0.4~mm from the estimated $35\times26$ depth map and calculates their centroid. For the latter two strategies, tactile observations are averaged over a 5-s window. The centered nine-frame input introduces a four-frame latency of approximately 0.13~s at a 30-Hz sensing rate. When both fingertips provide valid estimates, their lateral offsets are transformed into the same gripper coordinate direction and averaged; otherwise, the valid fingertip estimate is used. If the fused contact center lies outside the target band, the arm moves laterally toward the band and repeats the tactile measurement. No preset upper limit is imposed on the number of adjustments; the adjustment loop terminates when the estimated center enters the band. Because the hand uses separate joint configurations for contact-center estimation and lifting, with a larger fingertip separation in the lifting configuration, an off-center contact accepted during estimation may slip or fail to form a stable grasp after the configuration switch. The object is then lifted by 50~mm and held for 5~s, and the trial succeeds if it remains grasped throughout the holding period.


\subsection{Tactile-Guided Compensation}

Because visually tracked fingertip poses do not explicitly encode the local contact deformation in the human demonstration, the replay target is corrected along the fingertip normal, as shown in Fig.~\ref{S4_exp2_compensate_pipeline}. A scalar depth signal $d_i(t)$ is computed as the mean of the 10 largest values in the estimated tactile depth map, expressed in millimeters, and the corrected target is defined as
\begin{equation}
\mathbf{p}^{\mathrm{adj}}_{i}(t)=
\mathbf{p}^{v}_{i}(t)+\alpha_i d_i(t)\mathbf{n}_i(t),
\end{equation}
where $\mathbf{p}^{v}_{i}(t)$ and $\mathbf{p}^{\mathrm{adj}}_{i}(t)$ are the visual and compensated fingertip positions in meters, respectively, and $\mathbf{n}_i(t)$ is the unit normal pointing toward the object. The finger-dependent gain $\alpha_i$, expressed in $\mathrm{m/mm}$, converts tactile depth into a normal displacement; the gain vectors reported below follow thumb--index--middle order. It was empirically selected as the smallest value that enabled repeatable contact recovery in preliminary trials. The compensated positions are then used as inverse-kinematics targets, introducing only a local correction along the contact normal.

\begin{figure}[!t]
\centering
\includegraphics[width=3.5in]{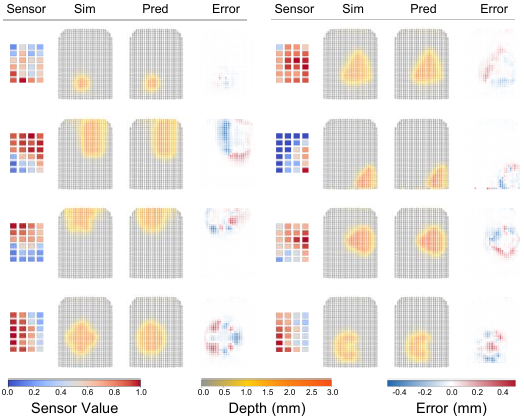}
\caption{Representative depth-map prediction results on test samples. For each sample, the sensor signals, simulation-generated depth label, predicted depth map, and error map are shown. }
\label{S4_exp1_examples}
\end{figure}

\begin{figure}
\centering
\includegraphics[width=\columnwidth]{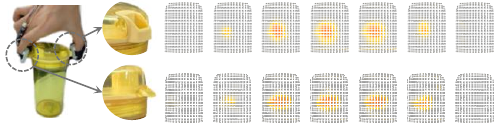}
\caption{TacPrint outputs during contact with a water bottle on two different surfaces. The thumb and index finger, each wearing a TacPrint, simultaneously press and release the flat surface and the edge of a water bottle, respectively. The resulting contact point-cloud sequences capture the characteristic contact patterns of the two surface types.}
\label{S4_exp1_real}
\end{figure}

\begin{table}[t]
\centering
\setlength{\tabcolsep}{2pt}
\caption{Prediction agreement with simulation-generated depth labels.}
\label{table_quant}
\begin{tabular}{clccc}
\toprule
Category & Metric & Median & P90 & Mean $\pm$ SD \\
\midrule
\multirow{3}{*}{Depth}
& Pixel-wise RMSE (mm) & 0.077 & 0.205 & $0.101 \pm 0.076$ \\
& Peak Depth Error (mm) & 0.095 & 0.484 & $0.188 \pm 0.258$ \\
& Contact-region RMSE (mm) & 0.172 & 0.446 & $0.223 \pm 0.161$ \\
\midrule
Position & Centroid Error (px) & 0.531 & 2.617 & $1.213 \pm 2.379$ \\
\midrule
Region & IoU & 0.883 & 0.949 & $0.829 \pm 0.169$ \\
\bottomrule
\end{tabular}
\end{table}


\subsection{Results and Analysis}
\textbf{Exp.~1: Depth-map estimation performance.}

Exp.~1 evaluates depth-map estimation from 24-channel capacitive inputs. As shown in Fig.~\ref{S4_exp1_examples}, the predictions generally agree with the simulation-generated labels in contact position and depth distribution, with most discrepancies occurring near contact boundaries and local peaks. A continuous-contact demonstration on a water bottle is shown in Fig.~\ref{S4_exp1_real}. 

Prediction agreement is evaluated using pixel-wise RMSE, peak-depth error, contact-region RMSE, depth-weighted centroid error, and intersection over union (IoU). Let $D^{\mathrm{pred}}$ and $D^{\mathrm{sim}}$ denote the predicted depth map and the corresponding simulation-generated label in millimeters, respectively, and let $P$ denote the total number of pixels. A pixel is considered in contact when its depth exceeds $\tau=0.1$~mm, giving the contact masks
$M^{q}=\{i\mid D_i^{q}>\tau\}$, where
$q\in\{\mathrm{pred},\mathrm{sim}\}$.
The global and contact-region depth errors are defined as
\begin{equation}
\begin{aligned}
\mathrm{RMSE}_{\mathrm{pix}}
&=
\sqrt{\frac{1}{P}
\sum_{i=1}^{P}
\left(D_i^{\mathrm{pred}}-D_i^{\mathrm{sim}}\right)^2},\\
\mathrm{RMSE}_{\mathrm{con}}
&=
\sqrt{\frac{1}{|M^{\mathrm{sim}}|}
\sum_{i\in M^{\mathrm{sim}}}
\left(D_i^{\mathrm{pred}}-D_i^{\mathrm{sim}}\right)^2}.
\end{aligned}
\end{equation}

The peak-depth error is
\begin{equation}
e_{\mathrm{peak}}
=
\left|
\max(D^{\mathrm{pred}})
-
\max(D^{\mathrm{sim}})
\right|.
\end{equation}

Contact localization is evaluated using depth-weighted centroids:
\begin{equation}
\begin{aligned}
\mathbf{c}^{q}
&=
\frac{\sum_{i\in M^{q}}D_i^{q}\mathbf{p}_i}
{\sum_{i\in M^{q}}D_i^{q}},
\qquad
q\in\{\mathrm{pred},\mathrm{sim}\},\\
e_{\mathrm{cent}}
&=
\left\|
\mathbf{c}^{\mathrm{pred}}
-
\mathbf{c}^{\mathrm{sim}}
\right\|_2,
\end{aligned}
\end{equation}
where $\mathbf{p}_i$ denotes the pixel coordinate. Finally, the contact-region overlap is measured by
\begin{equation}
\mathrm{IoU}
=
\frac{|M^{\mathrm{pred}}\cap M^{\mathrm{sim}}|}
{|M^{\mathrm{pred}}\cup M^{\mathrm{sim}}|}.
\end{equation}

Table~\ref{table_quant} shows agreement with the simulation-generated labels in depth, localization, and contact-region overlap, including a contact-region RMSE of $0.223\pm0.161$~mm, a weighted-centroid error of $1.213\pm2.379$ pixels, and an IoU of $0.829\pm0.169$.

With measured capacitive sequences as network inputs, the predicted depth evaluated at the guide-calibrated contact center showed a mean absolute error of $0.085\pm0.057$~mm across all 40 trials. This result evaluates the consistency of the predicted local indentation magnitude with the prescribed physical depth. Contact location was evaluated by comparing the centroid of the thresholded predicted region with the guide-calibrated contact center. For the 37 trials whose reference contact regions remained fully within the sensing area, the mean position error was $0.250\pm0.208$~mm, evaluating the localization consistency of the predicted contact region. The other three trials were retained for center-depth evaluation because the guide-calibrated center remained measurable, but were excluded from the primary position statistic because boundary truncation can bias an area centroid. For completeness, the position error across all 40 trials was $0.285\pm0.240$~mm.


\begin{figure}[!t]
\centering
\includegraphics[width=3.5in]{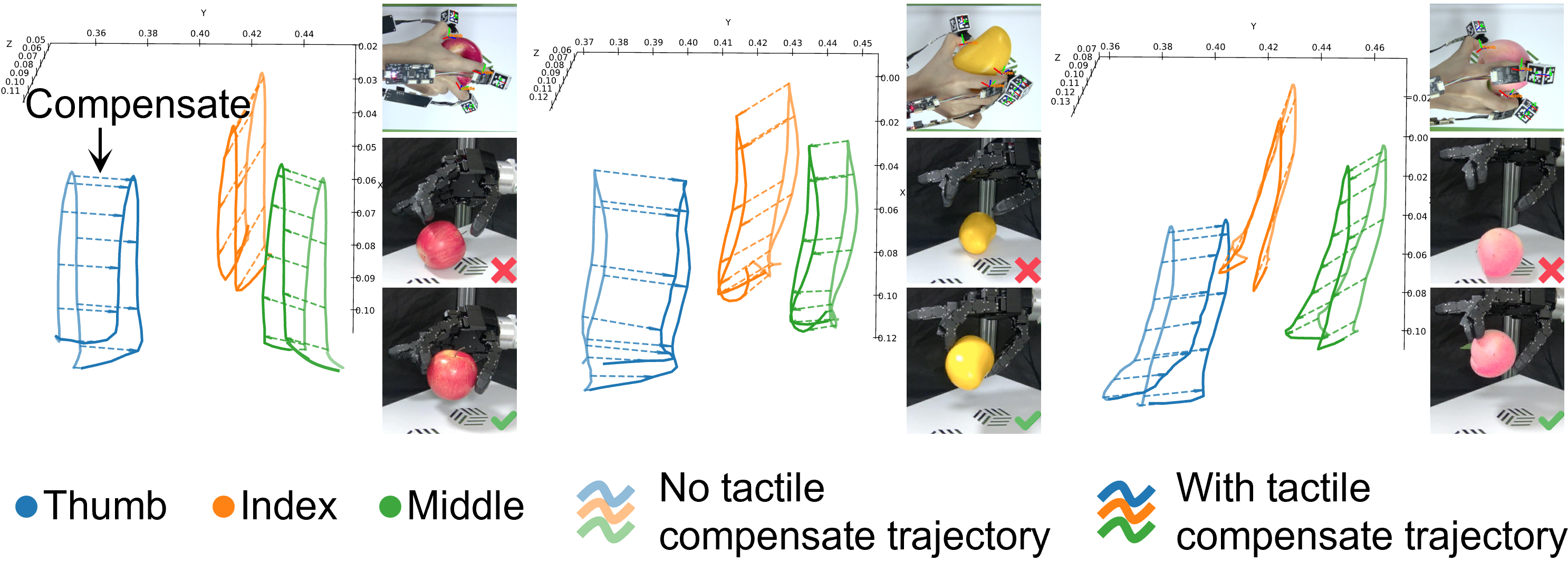}
\caption{Trajectory comparison between direct replay and replay with tactile-guided compensation of the grasping tasks. On the left of each grasping task, the transparent curves denote the direct replay trajectories, the solid curves denote the tactile-compensated trajectories, and the dashed arrows indicate the frame-wise displacement introduced by tactile compensation. On the right of each task, the first photo shows how the human demonstrator performs, the second shows that direct replay fails to complete the grasping task, and the third shows that with tactile compensation, the robot can complete the task.}
\label{S4_exp2_grasp}
\end{figure}

\textbf{Exp.~2: Tactile-informed robot replay.}
Exp.~2 evaluates whether TacPrint-derived tactile compensation can mitigate the contact mismatch encountered during vision-based robot replay. We compare direct replay using visually tracked fingertip trajectories~\cite{IKnyu} with replay augmented by normal-direction tactile compensation.

For fruit grasping, the gain vector was $\boldsymbol{\alpha}_{\mathrm{grasp}}=[0.005,0.005,0.005]$~m/mm. Direct replay positioned the fingertips close to the object surfaces but did not consistently recover sufficient normal contact for stable grasping. Tactile compensation introduced additional inward fingertip displacement and increased the success rates from 0\% to 95\%, 100\%, and 80\% for the apple, mango, and peach, respectively, yielding an overall success rate of 91.67\%, as shown in Fig.~\ref{S4_exp2_grasp}. 

\begin{figure}[!t]
\centering
\includegraphics{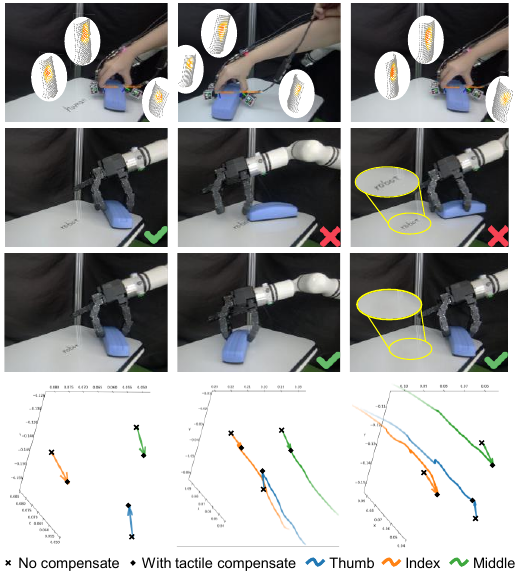}
\caption{Whiteboard-wiping experiments. The first row shows the human demonstration over two back-and-forth wiping passes and the tactile point clouds corresponding to the selected frames. The second row shows the corresponding direct replay, where the robotic hand fails to maintain the grasp and complete the task. The third row shows a replay with adjusted fingertip positions, where the robotic hand stably grasps the eraser and successfully removes the marker traces. The last row compares the fingertip positions before (crosses) and after (circles) tactile compensation, along with the compensated trajectories over several preceding frames. The compensated fingertips move further inward, resulting in a tighter grasp on the object.}
\label{S4_exp2_whiteboard}
\end{figure}

For whiteboard wiping, the gain vector was $\boldsymbol{\alpha}_{\mathrm{wipe}}=[0.004,0.010,0.010]$~m/mm. During direct replay, the fingertips did not establish a sufficiently tight grasp, and the eraser was lost during the first wiping stroke. Tactile compensation moved the fingertips further toward the eraser, maintained the grasp through two wiping passes, and increased the task success rate from 0\% to 90\%, as shown in Fig.~\ref{S4_exp2_whiteboard}. 

\begin{figure*}[!t]
\centering
\includegraphics[width=0.8\textwidth]{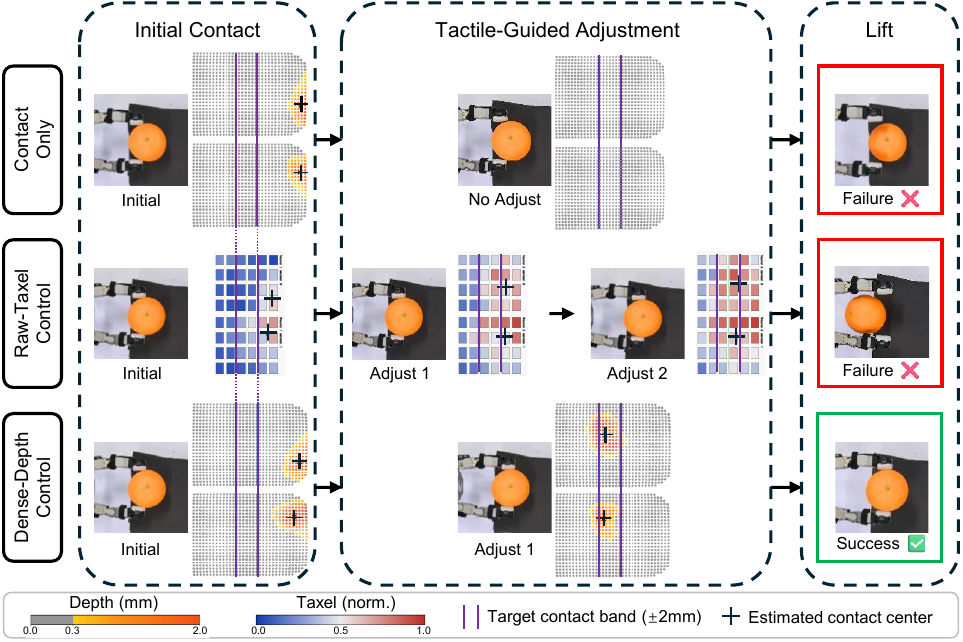}
\caption{Representative contact-position-aware grasping trials at Pos.~1. Contact only detects valid contact but performs no lateral adjustment and fails to lift the object. Raw-taxel control performs two lateral adjustments but still fails to lift the object. Dense-depth control performs one lateral adjustment and successfully lifts the object.}
\label{S4_exp3_paths}
\end{figure*}

\textbf{Exp.~3: Contact-position-aware grasping.}

Each strategy was evaluated in 40 trials, comprising five repetitions at each of the eight initial object positions. Among these, 20 trials corresponded to the four edge-contact positions, namely Pos.~1, Pos.~2, Pos.~7, and Pos.~8. Grasp success rate was used as the primary performance metric. For raw-taxel and dense-depth feedback, the number of lateral adjustments among successful trials was additionally reported to characterize the correction process. Because no preset adjustment limit was imposed, this number reflects the feedback strategy's own adjustment and stopping decisions rather than performance under a matched correction budget.

Figure~\ref{S4_exp3_paths} presents representative trials at Pos.~1. Contact-only feedback detected contact and immediately executed the grasp, but failed because the contact remained strongly offset. Raw-taxel feedback performed two lateral adjustments but still accepted an unstable grasping configuration. Dense-depth feedback moved the contact toward the predefined band after one adjustment and subsequently completed the lift. 

\begin{figure}[!t]
\centering
\includegraphics[width=\columnwidth]{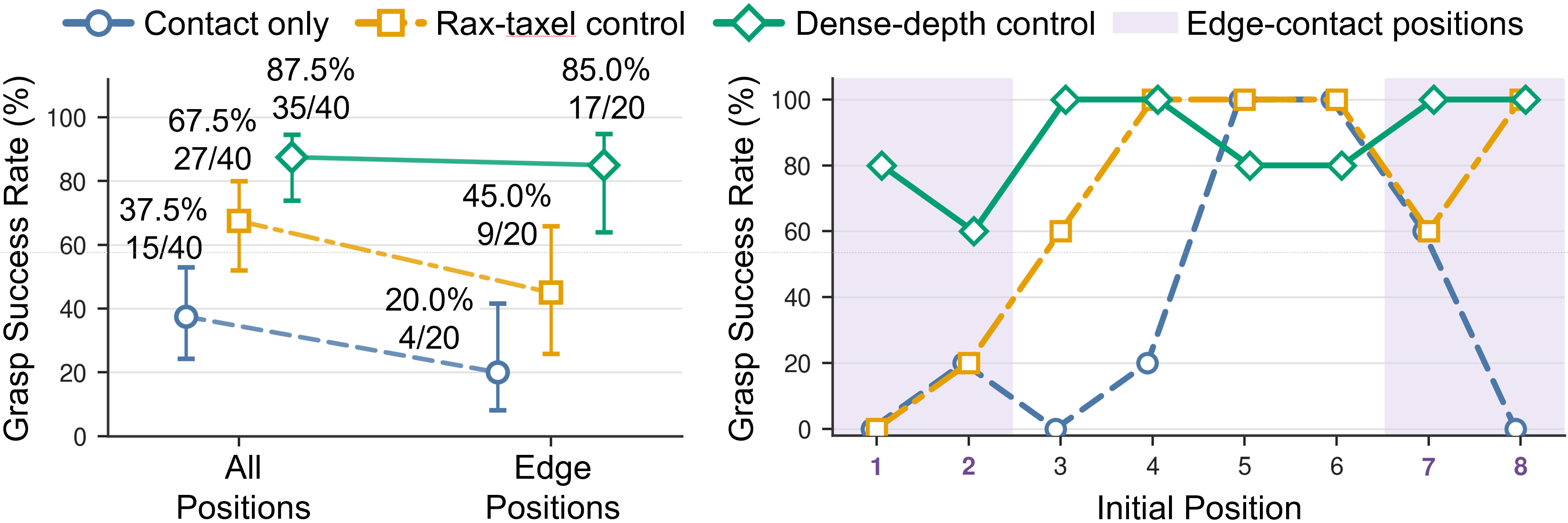}
\caption{Grasping performance of the three control methods. (a) Success rates over all positions and edge-contact positions, with 95\% Wilson confidence intervals. (b) Success rates at each initial position; shaded regions indicate edge-contact positions.}
\label{S4_exp3_results}
\end{figure}

As summarized in Fig.~\ref{S4_exp3_results}, contact-only, raw-taxel, and dense-depth feedback successfully completed 15/40, 27/40, and 35/40 trials, corresponding to success rates of 37.5\%, 67.5\%, and 87.5\%, respectively. Over the 20 edge-contact trials, the corresponding results were 4/20, 9/20, and 17/20, yielding success rates of 20\%, 45\%, and 85\%. Among successful trials, raw-taxel and dense-depth feedback required $1.07\pm1.36$ and $2.14\pm1.24$ lateral adjustments, respectively; under edge-contact conditions, the corresponding values were $2.56\pm1.01$ and $2.71\pm1.21$.

Contact-only sensing confirms contact occurrence but cannot determine whether the object lies within a suitable grasping region. Raw-taxel feedback provides coarse spatial information and improves the overall success rate, but the centroid of the sparse taxel responses can be biased toward the interior of the array, causing some off-center contacts to be accepted prematurely. The lower mean adjustment count of raw-taxel feedback therefore should not be interpreted as higher control efficiency: it was calculated only over successful trials and coincided with substantially lower overall and edge-contact success rates. Dense-depth feedback sometimes invoked additional corrections before reaching its stopping condition, but improved the overall success rate by 20 percentage points and the edge-contact success rate by 40 percentage points relative to raw-taxel feedback. The main benefit of the dense representation is therefore more reliable adjustment and stopping decisions rather than fewer corrective motions.

\section{Conclusion}

In this work, we presented TacPrint, a compact and low-cost wearable fingertip sensor with adjustable human mounting, taxel-aligned silicone protrusions, and customizable robotic connectors. A real-to-sim-to-real learning pipeline estimates a $35\times26$ contact-depth map from 24-channel capacitive signals. Against simulation-generated labels, the model achieved a contact-region RMSE of $0.223\pm0.161$~mm, a weighted-centroid error of $1.213\pm2.379$ pixels, and an IoU of $0.829\pm0.169$. With measured capacitive inputs, the network-predicted depth evaluated at the guide-calibrated contact center showed a mean absolute error of $0.085\pm0.057$~mm across all 40 trials. For the 37 trials whose reference contact regions were not truncated by the sensing boundary, the mean contact-position error was $0.250\pm0.208$~mm.

TacPrint supported two downstream robotic applications. Tactile-guided normal compensation increased grasping and whiteboard-wiping success rates from 0\% to 91.67\% and 90\%, respectively. In closed-loop contact-position-aware grasping, dense-depth feedback achieved success rates of 87.5\% overall and 85\% under edge-contact conditions, exceeding the corresponding raw-taxel results of 67.5\% and 45\%. Although dense-depth feedback did not reduce the mean number of adjustments among successful trials, its substantially higher success rates indicate more reliable contact-position estimation and stopping decisions, particularly near the sensing boundaries. These results support the use of the estimated depth representation for both human-to-robot replay and online grasp adjustment.

Limitations: The 24-channel input limits detailed boundary and shape reconstruction, and the predictions may retain characteristics of the training indenters. Moreover, the current physical evaluations validate contact depth and location rather than complete pixel-level deformation of the real elastomer. Residual discrepancies may therefore remain in contact boundaries and detailed local shapes, which should not be inferred solely from the reported center-depth and position errors. Future work will improve prediction accuracy, robustness across contact objects and wearing conditions, and deployment convenience.




\bibliographystyle{IEEEtran}
\bibliography{ref}

 




\vfill

\end{document}